\PassOptionsToPackage{table,usenames,dvipsnames}{xcolor}

\documentclass[authorversion,screen]{acmart}
\setcopyright{none}
\AtBeginDocument{%
  \providecommand\BibTeX{{%
    \normalfont B\kern-0.5em{\scshape i\kern-0.25em b}\kern-0.8em\TeX}}}

\usepackage{arydshln}
\usepackage{caption}
\usepackage{subcaption}
\usepackage{booktabs}
\usepackage{fontawesome}
\usepackage{multirow}
\usepackage{url}
\usepackage{pifont}

\author{Guilherme H. Resende}
\affiliation{%
  \institution{UFMG}\country{Brazil}
}

\author{Luiz F. Nery}
\affiliation{%
  \institution{UFMG}\country{Brazil}
}

\author{Fabrício Benevenuto}    
\affiliation{%
  \institution{UFMG}\country{Brazil}
}

\author{Savvas Zannettou}
\affiliation{%
  \institution{TU Delft}\country{Netherlands}
}

\author{Flavio Figueiredo}
\affiliation{%
  \institution{UFMG}\country{Brazil}
}

\begin{document}

\title{A Comprehensive View of the Biases of Toxicity and Sentiment Analysis\\ Methods Towards Utterances with African American English Expressions}

\renewcommand{\shortauthors}{Resende, et al.}

\begin{abstract}
Language is a dynamic aspect of our culture that changes when expressed in different technologies and/or communities. On the Internet, social networks have enabled the diffusion and evolution of different dialects, including African American English (AAE). However, this increased usage of different dialects is not without barriers. One particular barrier, the focus of this paper, is on how sentiment (Vader, TextBlob, and Flair) and toxicity (Google's Perspective and models from the open-source Detoxify) scoring methods present biases towards utterances with AAE expressions. Consider Google's Perspective (a toxicity scoring API) to understand bias. Here, an utterance such as ``All n*ggers deserve to die respectfully. The police murder us.'' it reaches a higher toxicity score than ``African-Americans deserve to die respectfully. The police murder us.''. This score difference likely arises because the tool cannot understand the re-appropriation of the term ``n*gger''. One explanation for this bias is that AI models are trained on limited datasets, and using such a term in training data is more likely to appear in a toxic utterance. While this may be a plausible explanation, the tool (if employed on a website) will make mistakes regardless of the explanation. Here, we study bias based on experiments performed on two Web-based (YouTube and Twitter) datasets and two spoken English, interview-based datasets. Our analysis shows how most models present biases towards AAE in most settings. We isolate the impact of AAE expression usage via linguistic control features from the Linguistic Inquiry and Word Count (LIWC) software, grammatical control features extracted via Part-of-Speech (PoS) tagging from Natural Language Processing (NLP) models, and the semantic of utterances by comparing sentence embeddings from recent language models. We present consistent results on how a heavy usage of AAE expressions may cause the speaker to be considered substantially more toxic than non-AAE speakers, even when speaking about nearly the same subject. Our study complements similar analyses focusing on small datasets and/or one method only. We are the first to compare six well-known methods, presenting explanations for scores with controls for grammar (PoS), linguistic features (LIWC), and semantics while employing four different datasets. Our results re-iterate that bias is still present in most methods and can also guide system developers in choosing the right tool for their use case. 
\end{abstract}




\received{22 February 2024}
\received[revised]{12 March 2024}
\received[accepted]{5 June 2024}

\maketitle

\section{Introduction}

In recent decades, we have witnessed a substantial rise in Internet usage. According to \cite{owidinternet}, Internet users increased from approximately 400 million in 2000 to 4.7 billion in 2020. With this increase in usage, it is natural that Web applications enable a wide diversity of social groups to interact among themselves and with other groups. Since such applications foster a more open and dynamic form of speech, a natural increase in the written form of dialects that previously were predominantly seen in the spoken form \cite{blodgett2016demographic} occurred. However, such massive amounts of textual data make manual content moderation impracticable. In other words, the heavy usage of social media has evidenced the urge for automatic moderation tools that measure and moderate improper behavior online. One of the main concerns is the public display of negative/toxic sentiments against a person or specific group, more drastically when the target is a minority group historically marked with discrimination and stereotypes. The necessity of dealing with the increasing number of deviating content has led many researchers and companies to use AI tools to identify such events \cite{ribeiro2016sentibench}.

Concurrently to the increase in Web usage, African-American English (AAE) has gone from being seen as a marginalized dialect of English to a consolidated vernacular of the language~\cite{green2002african}. Like most dialects, the AAE was initially heavily used in spoken form and had the Web as a crucial influence on its emergence in the written form \cite{blodgett2016demographic}. However, and as we have discussed, the Web is not only a disseminator of cultural aspects of our society but also a vehicle where toxicity campaigns against African Americans are prone to occur\footnote{\url{https://theconversation.com/the-rings-of-power-is-suffering-a-racist-backlash-for-casting-actors-of-colour-but-tolkiens-work-has-always-attracted-white-supremacists-189963}}. Even though several websites have well-defined community guidelines, user anonymity and lack of unaccountability leave room for misbehavior. 

\begin{table}[t!]
\scriptsize
\centering
\begin{tabular}{p{1cm}lrrrrrr} \toprule
& {\bf Score} & \multicolumn{1}{p{1.9cm}}{\texttt{All my friends on the porch and never in the house}} & \multicolumn{1}{p{1.9cm}}{\texttt{All my n*ggas on the porch and neva ina house}} & \multicolumn{1}{p{1.3cm}}{\texttt{You're white}} & \multicolumn{1}{p{1.3cm}}{\texttt{You're black}} & \multicolumn{1}{p{1.7cm}}{\texttt{I can't forget you}} & \multicolumn{1}{p{1.7cm}}{\texttt{Cant fuhgit you}} \\ \midrule
Persp. (ML) & [\colorbox{SpringGreen}{\faThumbsOUp=0}, \colorbox{OrangeRed}{\faThumbsODown=1}] & \cellcolor{Goldenrod} 0.2396 & \cellcolor{OrangeRed} 0.7886 & \cellcolor{Goldenrod} 0.2546 &  \cellcolor{Orange} 0.4256 & \cellcolor{SpringGreen} 0.0406 & \cellcolor{Orange} 0.2359 \\
Detox (ML) & [\colorbox{SpringGreen}{\faThumbsOUp=0}, \colorbox{OrangeRed}{\faThumbsODown=1}] & \cellcolor{SpringGreen} 0.0012 & \cellcolor{Orange} 0.6145 & \cellcolor{OrangeRed} 0.8766 & \cellcolor{OrangeRed} 0.9718 & \cellcolor{SpringGreen}0.0257 & \cellcolor{OrangeRed} 0.7601 \\
Detox U & [\colorbox{SpringGreen}{\faThumbsOUp=0}, \colorbox{OrangeRed}{\faThumbsODown=1}] & \cellcolor{SpringGreen} 0.0013 & \cellcolor{Orange} 0.6842 & \cellcolor{Goldenrod} 0.2549 & \cellcolor{OrangeRed} 0.8903 & \cellcolor{SpringGreen} 0.0162 & \cellcolor{Orange} 0.5332 \\
\midrule
Vader (L) & [\colorbox{SpringGreen}{\faThumbsOUp=0}, \colorbox{OrangeRed}{\faThumbsODown=1}] & \cellcolor{SpringGreen}0 & \cellcolor{SpringGreen}0 & \cellcolor{SpringGreen}0 & \cellcolor{SpringGreen}0 & \cellcolor{SpringGreen}0 & \cellcolor{SpringGreen}0 \\
Textblob (L) & [\colorbox{OrangeRed}{\faThumbsODown=-1}, \colorbox{SpringGreen}{\faThumbsOUp=1}] & 0 & 0 & 0 & \cellcolor{SpringGreen} -0.1666 & 0 & 0 \\
Flair (ML) & [\colorbox{OrangeRed}{\faThumbsODown=-1}, \colorbox{SpringGreen}{\faThumbsOUp=1}] & \cellcolor{SpringGreen} 0.9830 & \cellcolor{OrangeRed} -0.7296 & \cellcolor{SpringGreen} 0.9994 & \cellcolor{SpringGreen} 0.9992 & \cellcolor{SpringGreen} 0.9957 & \cellcolor{SpringGreen} 0.9111 \\ \bottomrule
\end{tabular}
\caption{Comparing the Toxicity Scoring Models -- Perspective, Detoxify, and Detoxify Unbiased -- as well as the Sentiment Scoring -- Vader, Textblob, and Flair -- in six utterances. Utterances are paired based on meaning. Good scores (\colorbox{SpringGreen}{\faThumbsOUp}) are colored green, bad scores (\colorbox{OrangeRed}{\faThumbsODown}) are colored red. Other shades represent scores in between. For Textblob and Flair, \colorbox{OrangeRed}{\faThumbsODown=-1} are utterances of negative sentiment. In Perspective, Detoxify, and Vader, a negative sentiment or toxic utterance has a score of \colorbox{OrangeRed}{\faThumbsODown=1}. (ML) indicates a Machine Learning model, whereas (L) indicates a lexical or rule-based approach. On this table, lexical approaches are less biased.}
\label{tab:examples}
\end{table}

The aforementioned rise in AI moderation tools (such as Google's Perspective~\cite{perspective} and others~\cite{ribeiro2016sentibench,giachanou2016like,yadollahi2017current}) aim to reduce the amount of negative or toxic utterances online. Overall, such tools rely on Machine Learning (ML) models that help determine proper and improper utterances. Nevertheless, as previous research has discussed, automatic content moderation can backfire and present biases towards minorities~\cite{sap2019risk,hutchinson2020unintended,venkit2021identification,bolukbasi2016man}. For instance, a tool for toxicity analysis may present high scores for non-toxic AAE sentences for no apparent reason. To depict this issue, we show examples of toxicity and sentiment analysis models employed in online text. We point out that it is quite easy to find problematic utterances when we employ slang terms such as {\em ``n*ggas''}. In Table \ref{tab:examples}, we contrast three pairs of sentences that should reach similar toxicity/negative sentiment levels.

\textit{Why does the problem arise?} From a linguistic perspective, dialects may inherently manifest behaviors and cultural aspects of the groups in which they were created \cite{ball1992cultural,florini2014tweets,friedrich2016sociolinguistics,friedrich2020englishes}. Terms such as ``n*gger'' are problematic for AI models since both the term and its variations have a historical pejorative usage\footnote{\url{https://en.wikipedia.org/wiki/Nigga}}. Nevertheless, this same term was re-appropriated by the black community so that its use ceased to be considered problematic when used by people inside the black community. Suppose such a fine line between causal speaking and offensive discourse is problematic from a human and computational perspective. In that case, these interpretations are confounding to automatic content moderation tools. In other words, toxicity/sentiment analysis tools are usually developed using manual rules or supervised ML techniques that employ human-labeled data to extract patterns. The disparate treatment embodied by machine learning models usually replicates discrimination patterns historically practiced by humans when interacting with processes in the real world. Due to biases in this process, a lack of context leads both rule-based and machine learning-based models to a concerning scenario where minorities do not receive equal treatment~\cite{gonen2019lipstick,tatman2017gender,abid2021large,diaz2018addressing}.


This discussion leads to the research question behind our paper: \textit{Is there a systematic bias on toxicity/sentiment analysis towards AAE?} To better understand this issue, we present a broad-scale analysis. To do so, we manually curate a dataset of African American English Expressions~\cite{smitherman2000black,widawski2015african} (these sources were recommended by the organizers of the to-be-released Oxford Dictionary of African American English\footnote{\url{https://hutchinscenter.fas.harvard.edu/odaae}}). These expressions complement four different utterance datasets with some demographic information on race (i.e., interviews from African American individuals, AAE utterance vs non-AAE utterance labels, and author-supplied labels). {\em We emphasize that we cannot state how a speaker identifies regarding her/his race for some datasets. AAE may also be employed by non-African Americans. However, we interpret our results using AAE expressions from our expressions dataset (made available)}.

The models we study can be divided into toxicity (Google's Perspective~\cite{perspective}, Detoxify, and Detoxify Unbiased~\cite{detoxify}), and sentiment analysis (Flair~\cite{akbik2019flair}, TextBlob~\cite{loria2018textblob}, and Vader~\cite{hutto2014vader}) models, but also can be segmented into machine learning-based (Google's Perspective, Flair, Detoxify, and Detoxify Unbiased), and lexical, or rule, based (Vader, and TextBlob) models. Our main contributions are as follows:
\begin{enumerate}
\item We present a broad-scale analysis of biases toward utterances with AAE expressions in six out-of-the-shelf, well-known models and/or APIs;
\item To do so, we focus on unveiling if there is a systematic tendency for AAE utterances to be deemed more toxic or negative sentiment by several models of datasets of different natures (tweets, closed captions, and spoken interviews). To reach our results, we manually transcribed a dictionary of AAE expressions and used the number of such expressions in an utterance as an explanatory feature;
\item Other control features include Lexical Analysis~\cite{pennebaker2001linguistic}, and grammar-based Part of Speech Tagging PoS) labels for words in utterances. Overall, we discuss which characteristics of the utterance lead the model to deem it as toxic or of negative sentiment. The number of AAE expressions is a recurrent statistically significant feature;
\item Using recent language models~\cite{wang2020minilm}, we show that in some datasets, even utterances from African-American (AA) speakers that have a similar meaning to those from non-AA speakers, models will, in several cases, score the sentence from non-Whites with more toxic or more negative sentiment scores.
\end{enumerate}

Before continuing, we point out that our work is not the first to study the biases of similar models towards minorities~\cite{gonen2019lipstick,tatman2017gender,abid2021large,diaz2018addressing}. However, we complement prior endeavors with broader-scale analysis. Previous methods focus on a single dataset or model and do not perform the lexical and grammar-based analysis we do here. We further point out that our goal in this paper is not to pinpoint models with the best accuracy. We focus on showing AAEs and comparing if there is a tendency across several models. The datasets where we show this issue range from online texts from Twitter~\cite{blodgett2016demographic,blodgett2018twitter}, spoken English datasets gathered by linguists \cite{pitt2005buckeye,kendall2021corpus}, and online single speaker English from YouTube movie reviews. The YouTube dataset (see Section \ref{sec:data}) was a manual effort toward gathering data with fewer confounding factors (i.e., single-speaker videos). This dataset is made available to the community to improve the current and yet-to-come NLP models\footnote{\url{https://anonymous.4open.science/r/aae_bias-D396/data/aae_terms_black_talk.yaml}}.

Our results show that biases are more prominent on online datasets, such as Twitter and YouTube, and less strongly but still present in spoken English interviews. Our research shows that using AAE expressions will likely lead to sentences being deemed more toxic, even when sentences are similar to those with non-AAE expressions. Overall, system developers may use these findings to determine what model type shall be employed (sentiment analysis vs. toxicity scoring) or whether ML vs. lexical-based models are more adequate for their application. More importantly, our findings show that even considering ``unbiased models''~\cite{detoxify}, ML models still present a bias towards utterances with AAE expressions. Indicating that AAE speakers may still face unwarranted moderation online. 

The next section presents background knowledge and previous research related to our effort. This will be followed by our data description in Section~\ref{sec:data}. Section~\ref{sec:results} presents our results and Section~\ref{sec:conc} our concluding discussion.


\section{Background and Related Work}

This section presents an overview of the literature related to our study. We begin discussing the sociolinguistics of English online, as the increased usage of AAE expressions online is a primary motivation behind our work. Subsequently, we also discuss the motivation behind sentiment analysis tools, available alternatives, their major strengths and shortcomings, and how toxicity relates to sentiment analysis. Next, we discuss bias in machine learning methods and how they can negatively influence individuals online and suppress the discourse of minority groups. Finally, we focus on those papers most related to ours and present a statement on the novelty of our research.

\subsection{Sociolinguistics of AAE and Other Englishes Online}

As a research field, sociolinguistics focuses on studying how social context affects the usage and evolution of language. Overall, humans take part in several speech communities throughout their lives~\cite{friedrich2016sociolinguistics}, and even the same human being may communicate in different variations of English depending on the community she/he is interacting with. With the rise of the Social Web over the 2000's and 2010's, the field also focused on how Web communities affect language~\cite{friedrich2016sociolinguistics,friedrich2020englishes}. In particular, Friedrich and Figueiredo~\cite{friedrich2016sociolinguistics} argue that hundreds of years after the invention of the printing press, the written usage of English as a language appeared to be ``evolving'' to a standard or uniform English. However, with the Web, community and individualized language use has increased over recent years. AAE is an example of such a case~\cite{friedrich2016sociolinguistics,friedrich2020englishes,blodgett2016demographic}, where the dialect has experienced a rise in usage (particularly online) in recent years.

One example of the expansion of AAE in recent years comes from the movement known as {\em signifyin’}~\cite{florini2014tweets}. In other words, when {\em signifyin’} one expresses their race via particular dialects, such as AAE, on social media. This expression is utilized to resist the oppression present in one's day lives~\cite{florini2014tweets,nakamura2013cybertypes}. Regarding how AAE is spread online, some authors argue that the dialect spreads initially from Web users from large cities to smaller communities in wave-like, or viral, patterns~\cite{eisenstein2014diffusion}. Overall, Frierich and Figueiredo state that:

``With the Internet, we have witnessed a change in this scenario. Gender and racial/ethnic activism have become quite strong online and have served not only to spread the debates but also to add new layers to them – such as the complex construction of identities in cyberspace. And again, we must say, English has been quite present in this new picture, mainly because of its lingua franca status and association with technology.''

Our debate so far focuses on how the Web helped to propagate AAE. However, for several decades, the dialect has also been studied offline~\cite {widawski2015african}. Over the years, some attempts have been made to catalog different expressions from the dialect~\cite{dillard1977lexicon,baugh1981runnin,smitherman2000black}. A partnership between Oxford and Harvard also organizes a yet-to-be-released dictionary called the Oxford Dictionary of African American English (ODAEE).

Given that language constantly evolves, we aimed to collect a recent corpus of AAE expressions. To do so, we contacted the organizers of the ODAAE, asking if they could share the list of expressions used in their dictionary. The organizers kindly denied it because ODAEE is still a work in progress. Nevertheless, they did suggest that we use expressions from Smitherman's Black Talk Dictionary~\cite{smitherman2000black} as it is a large and somewhat recent corpus. In our research, we manually transcribed every expression from this dictionary as our AAE expression list.

\subsection{Sentiment Analysis and Toxicity Models}
\label{section:rw_sentiment_analysis}
Sentiment Analysis identifies sentiments and quantifies their intensity (positive or negative sentiment) in utterances. Current Sentiment Analysis models may be classified into two major categories, namely, machine learning-based (ML models) and lexical-based approaches, described below. These methods have been employed since the mid-2000s, and one of the major motivations behind building such approaches was to rate user reviews online.

ML-based Sentiment Analysis models \cite{go2009twitter,wang2012system,yang2018feature,song2018boosting,akbik2019flair} are built over a sample of data points comprising as many examples as possible from positive and negative sentences. Usually, the learning procedure targets data drawn from a context of interest (e.g., Twitter, Facebook, Marketplaces, etc.), and humans manually label this dataset to train models. This family of methods often benefits from complex word representations and can grasp deeper relationships implied in daily conversations. Lexical-based approaches \cite{pennebaker2001linguistic,hu2004mining,taboada2011lexicon,hutto2014vader}, on the other hand, begin by listing seed words considered to be representatives of groups of emotions. Once the seed list is complete, it is incremented with similar words and synonyms. Such approaches must actively deal with normal word usage that may change the intent/intensity of the sentence, such as negations, punctuations, capitalization, etc. Since this approach is based on the human understanding/application of terms and expressions, their performance on novel datasets may have less statistical variance (e.g., don't overfit).

Compared to lexical-based methods, the ML models rely on the vector representation of terms and utterances \cite{mikolov2013distributed}. Such vector representations are used as inputs to train supervised methods. Collecting high-quality labels to train such models is a difficult-to-reach pre-requisite (discussed below). On the other hand, lexical-based approaches need to explicitly address negations, punctuations, out-of-vocabulary occurrences, and more complex relations between terms \cite{ribeiro2016sentibench}. To address the gaps left by each family of methods, authors have also proposed hybrid solutions \cite{pappas2013distinguishing,thelwall2014heart,wilson2005opinionfinder}. 

More recently, we have seen a rise in Toxicity Classification (compared to Sentiment Analysis) models \cite{koratana2018toxic,perspective,detoxify}. Most, if not all, of these approaches are ML-based. Toxic speech is usually considered to be an umbrella term that comprises hate speech, abusive language, racism, and so on \cite{harris2022exploring, kumar2021designing}. Despite the efforts to address toxic speech, there is not a clear agreement about what it means for a sentence to be toxic. One of the most established definitions is presented by \cite{dixon2018measuring}, which defines toxicity as rude, disrespectful, or unreasonable language.

Due to the lack of consensus on toxicity, the inherent ambiguity of labeling sentences presents an issue to ML models. The vast majority of datasets use human labelers, which are influenced by their previous experiences and, most of the time, do not have access to the underlying context from which the respective sentence was drawn. This subjectivity and lack of context may cause considerable labeling issues. For example, Kumar et al. \cite{kumar2021designing} state that people who have suffered harassment in the past are more prone to label random sentences from some social networks as toxic than those who did not face such problems. Maybe due to its less restrictive definition and to the capacity to encompass many types of harassment, toxicity models are actively used in practice to moderate discourse in many platforms~\cite{perspective-el-pais, perspective-le-monde, perspective-nyt}; however, with some known bias problems.

\subsection{Biases Towards Minorities Online}

We now discuss prior work on the biases of Web datasets and AI models. Starting from Jia et al.~\cite{jia2015measuring}, the authors investigated the proportions in which men and women appeared in news articles' images. The authors found that men are considerably more frequently represented than women. Garcia et al.~\cite{garcia2014gender} also described a consistent bias towards men in Twitter content. On Twitter, female users tend to describe more events in which men play important roles. Babaeianjelodar et al.~\cite{babaeianjelodar2020quantifying} explored the nuances of gender biases over ML models. In all datasets considered, models perform disparately against unprivileged subgroups. Similar findings were raised by several other authors considering countries~\cite{graham2014uneven}, age~\cite{diaz2018addressing},  religion~\cite{abid2021large}, and sexual orientation ~\cite{dixon2018measuring}. Regarding dialects, Blodgett et al.~\cite{blodgett2016demographic} studied how language characteristics can change considerably within the same country. The work focuses on learning distinguishable features between Mainstream American English (MAE) and AAE with a geographic context. In \cite{gomes2019drag}, the authors also present another clear differentiation between English focusing on Drag queens. Here, the authors find that transgender individuals have a speaking characteristic consistently seen as more toxic by ML models. 

As Bamman et al.~\cite{bamman2014distributed} states, language is always situated within a context. Neglecting this surrounding context leads to disparate treatments. For example, transgender individuals may be speaking a dialect deemed toxic if used by someone outside of the community. However, this may be a defense mechanism to cope with tough situations imposed by society \cite{gomes2019drag}. Similar language signals are passive within the black community and the AAE dialect. Studies were already performed to comprehend and measure how much ML models are biased against AAE speakers \cite{blodgett2016demographic,ball2021differential,sap2019risk}.

Overall, we can state that nowadays, it is not hard to find discrimination episodes involving AI models~\cite{diaz2018addressing,abid2021large,dixon2018measuring,hutchinson2020unintended}. For example, Abid et al.~\cite{abid2021large} interacted with a conversational artificial intelligence model touching religion-related subjects and noting the inner associations with the topic. Finally, they found a consistent bias associating Muslims with terrorists (in 23\% of the test cases) and the Jewish with money (in 5\% of the test cases). In the opposite direction, mitigating such biases are also common~\cite{bolukbasi2016man,dixon2018measuring,blodgett2016demographic,babaeianjelodar2020quantifying,harris2022exploring}. Nevertheless, as studied by Gonen et al.~\cite{gonen2019lipstick}, persistent bias may stick with the model even after active effort has been applied to remove it. We also observe this as we use unbiased versions of recent models. Since the ML model complexity has increased in the last few years, we could expect the bias to be more elaborate and complex to fight against. This leads us to the problem of using biased models for sensible tasks that may perpetrate harmful behavior. Currently, sentiment and toxicity analysis models are deliberately used to moderate forum discussions of relevant news media and magazines \cite{perspective-nyt,perspective-le-monde,perspective-taringa,perspective-el-pais}.

\subsection{The Literature on Bias of Toxicity/Sentiment Scoring Models and Research Novelty}

We now detail prior work that is most related to ours (e.g., evaluating and unveiling biases in similar models)~\cite{wright2021recast,muralikumar2023human,hosseini2017deceiving,grondahl2018all,field2021survey,kiritchenko2018examining,sap2019risk,davidson2017automated,soremekun2022astraea}. Before doing so, we initially point out that the studying the biases of NLP models towards race is a well-established research topic and the survey of Field et al.~\cite{field2021survey} presents a recent overview on this topic. The authors of this survey analyzed 79 different papers on race and NLP systems. Overall, the consensus is that NLP still encodes racial biases (something we also observe) and that race is commonly studied as a limited categorical variable. Here, we take a step towards a broader view of the issues by incorporating in our study (1) a novel list of AAE expressions, (2) grammatical features (PoS), and (3) linguistic features (LIWC) to understand biases.  

One of the vanguard efforts of looking into the biases of toxic scoring APIs (Perspective in particular) was performed by Sap et al.~\cite{sap2019risk}. In contrast, Kiritchenko et al. were among the first to study the bias of sentiment analysis models~\cite{kiritchenko2018examining}. Starting with the former, the authors discuss how datasets are biased and how models propagate such biases. However, unlike our work, the authors only study Twitter datasets and do not present statistical analysis on how utterance features (grammar, linguistics, and usage of AAE expressions) may explain biases. The authors also only focus on Perspective as an out-of-the-shelf model. The former, focused on sentiment analysis, compares over two hundred models from a well-established information retrieval sentiment classification challenge. However, the author does not use real-world datasets as we do and focuses their analysis on sentences of similar meaning but with small changes in words related to gender, occupation, and race. This approach is similar to the adversarial attacks described next.

Although not focused on measuring biases, the work of Hosseini et al. ~\cite{hosseini2017deceiving}, and Gröndahl et al.~\cite{grondahl2018all} both show how small changes to a sentence will change model scores. We present a different view on this finding by incorporating semantic similarity using language models~\cite{wang2020minilm} and finding that even expressions that are semantically close to one another will differ in scores depending on the number of AAE expressions used. Similarly, Davidson et al.~\cite{davidson2017automated} discussed some challenges in differentiating hate speech from other offenses. This provides evidence of how language is nuanced, and models have problems with these small nuances. Wright et al.~\cite{wright2021recast} provides a tool called RECAST that helps users pinpoint words that need changing in order to reduce the toxicity of a score.

Regarding human evaluation of models, we refer to the recent effort of Muralikumar et al.~\cite{muralikumar2023human}. Here, the authors evaluate Perspective and contrast how human scores align with the model. Overall, the score from Perspective is a good predictor (based on a Logistic regression) of human labels (``toxic'', ``hard to say'', ``non-toxic''). However, agreement is not always present, with the model still making mistakes. The authors suggest that using a score cut-off of 0.55, i.e. if the model scores over this value classify the utterance as toxic, will make model outcomes agree with users 50\%. 

Testing different definitions of fairness is also an active field of research~\cite{chen2022fairness}, with software tools being developed just for this task in NLP models~\cite{soremekun2022astraea}. We complement these efforts by showing that out-of-the-shelf models and API still require further testing and mitigations, as biases towards utterances with AAE expressions are still quite present.








Our research differs from previous works by investigating biases in models of different families (ML-based and lexical-based methods) and throughout four datasets representing different contexts (in-person conversations, single-speaker movie reviews, and personal social media posts). Here, we focus on out-of-the-shelve methods and those already applied in real-world forums. Unlike prior work, we employ different statistical features of utterances to show how the presence of AAE expressions will lead models to rate sentences as more toxic or of negative sentiment. Our statistical analysis also provides insights into what features models explore to reach their scores. Finally, the datasets we employed were collected to isolate confounding factors. That is: (1) we do not use any dataset used to train models or with toxicity/sentiment labels; (2) one of our datasets is focused on single-speaker movie reviews, controlling for discourse as a confounder; (3) we also compare models on well-established linguistic datasets on single-person interviews.


\section{Datasets and Pre-Processing} \label{sec:data}

We explore four datasets of different natures to understand the extent of biases in toxicity/sentiment analysis models and when they present themselves more strongly. Initially, we use the Twitter AAE dataset~\cite{blodgett2016demographic,blodgett2018twitter}. This dataset is interesting as it contains tweets classified as AAE or Mainstream American English (MAE). Tweets were classified using an ML model, and we consider a subset of tweets where the model predicted over 80\% probability for each class. This is a well-established dataset for AAE utterances used by other endeavors~\cite{sap2019risk}. Twitter is one of the major platforms where one would expect that toxicity and sentiment analysis models could mitigate unwanted behavior. On the negative side, as the dataset contains general Tweets, it does not control for confounding factors such as dialogues, debates, and potentially controversial topics. Thus, we complement this research with two other datasets described next.

Our YouTube dataset comprises subtitles extracted from YouTube movie reviews with a single speaker discoursing about a unique topic per video. The topics are movies from Rotten Tomatoes 100 Best Movies of All Time. We targeted single-speaker videos to control for any confounding variables that may appear with dialogues. Also, we focus on acclaimed films\footnote{\url{https://www.rottentomatoes.com/top/bestofrt/}} to control for the possible negative influence of bad content (speakers may still dislike the movies, though). Our goal with the YouTube dataset is to control both content and dialogue. 

Finally, we explore the linguistic Corpus of Regional African American Language (CORAAL) \cite{kendall2021corpus} as representations of spoken African-American English. For comparisons, we employ the Buckeye \cite{pitt2005buckeye} dataset focused on Caucasian\footnote{\url{https://buckeyecorpus.osu.edu/php/corpusInfo.php}} speakers from central Ohio. Both datasets are focused on spoken interviews that have textual transcripts. Buckeye was recommended to us by the curators of CORAAL.

Before detailing our datasets, we discuss how we identified African-American English expressions (AAE expressions). As stated, we manually transcribed the Black Talk dictionary\cite{smitherman2000black}. Since AAE first emerged as an oral language, the main intent of this dictionary was not to define the etymological history of terms; instead, it concentrates on the meanings and significance of expressions. The organizers of the not-yet-published Oxford Dictionary of African American English referred us to the Black Talk dictionary as a reliable source.

The Black Talk dictionary comprises more than 1800 entries. Since some entries are sentences instead of single terms, they may apply to different pronouns. In such cases, the possible use cases are listed. For example, ``BREAK HIM/HER/THEM OFF SOMETHING'' becomes three expressions. Our transcription of the entries considers every possible combination presented. The entire list of expressions is available for download.

\begin{table}[t!]
\scriptsize
\centering
\begin{tabular}{llrrrr}
\toprule
\textbf{Dataset} & \textbf{Demographic} & \textbf{\# Documents} & \textbf{\# Sentences} & \textbf{\# AAE Expr.} & \textbf{AAE Expr. per Doc.} \\ \midrule
\multirow{2}{*}{YouTube} & AA Speaker$\,^{\ast}$ & 150 & 17828 & 18308 & 122.05 (\rotatebox{90}{\color{green!80!black}\ding{225}} 43\%) \\ 
 & non AA Speaker$\,^{\ast}$ & 484 & 41464 & 42729 & 85.67 \\ \midrule
\multirow{2}{*}{Twitter} & AAE Tweet & 250 & 250 & 372 & 1.49 (\rotatebox{90}{\color{green!80!black}\ding{225}} 43\%) \\ 
 & MAE Teet & 250 & 250 & 259 & 1.04 \\ \midrule
CORAAL & AA & 142 & 64493 & 61651 & 434.16 (\rotatebox[origin=c]{270}{\color{red!80!black}\ding{225}} 9\%) \\
Buckeye & Caucasian & 39 & 19304 & 18712 & 479.79 \\ \bottomrule
\end{tabular}
\caption{Datasets statistics. The \textit{AAE Expressions Ratio} represents the average number of African-American English terms per document in the corpus. AA stands for African-American, AAE for African-American English, and MAE Mainstream American English. $\,^{\ast}$ Indicates that two (agreeing) authors inferred the demographic. Still, it is not used as a variable in our analysis (we rely on the \# of AAE Expressions as an explanatory variable in this case). \rotatebox{90}{\color{green!80!black}\ding{225}} and \rotatebox[origin=c]{270}{\color{red!80!black}\ding{225}} indicates the percentual increase/decrease in AAE expressions when comparing demographics. Surprisingly, CORAAL has a {\bf small} increase in AAE expressions.\vspace{-1em}}
\label{tab:data}
\end{table}

In Table~\ref{tab:data}, we present a summary of our datasets in the number of sentences (or utterances), number of words (non-unique), and number of African-American English expressions present. Over the next few subsections, we dive into the details of each dataset and how they were gathered.

{\bf Twitter:} The Twitter dataset comes from the Twitter AAE\footnote{\url{http://slanglab.cs.umass.edu/TwitterAAE/}} website. To create the dataset, the authors ~\cite{blodgett2016demographic,blodgett2018twitter} developed a Latent Dirichlet Allocation (LDA) based topic model that took into account both the frequency of common terms used in AAE as well as Census data. Based on the location where the account tweeted from, an initial race estimate is thus performed. This information is combined with the presence of key terms to derive different latent topics for the corpus. These topics were then explored to label AAE and non-AAE tweets.

Although the authors label over $80,000$ tweets, we focus on a smaller sample of $500$ tweets that the authors manually inspected. These tweets were manually labeled with PoS tags to derive an African-American English PoS model. According to the authors, more than 18\% of the terms used within the African-American tweets are not in the standard English dictionary. It is also very common to find words written in their phonological style in AAE - e.g. tha (the), iont (I don't), ova (over), and so on - while the contrary was found to never happen in the Non-AAE tweets. 

{\bf YouTube:} The YouTube dataset is a collection of subtitles from YouTube movie review videos. A single speaker talks to the audience about a movie production listed among the most relevant movies of all time. We considered Rotten Tomatoes' top 100 best movies of all time ranking due to their prestige among the audience and because they have a higher probability of being well-spoken in a review. For each of the top 50 movies from the ranking, two authors manually searched and cataloged as many videos as possible. The authors determined Demographic labels, namely, African-American Women, African-American Men, non-African-American Women, and non-African-American Men, in order of appearance when querying the movie name on YouTube. Since YouTube doesn't naturally disclose demographic information about its users, we had to restrict our search only to producers who happened to appear on the screen at least once throughout the entire video. The list of movies and the respective YouTube channels used in constructing this dataset is available~\footnote{\url{https://anonymous.4open.science/r/aae_bias-D396/data/youtube_data_description.csv}}.

When publishing videos on YouTube, the creators can either explicitly inform their videos' subtitles or let the YouTube transcription model automatically caption them. Nonetheless, differently from manually informed subtitles, the captioning mechanism, by default\footnote{\url{https://support.google.com/youtube/thread/70343381}}. For fair comparisons, we make use of automatic transcriptions as these are the ones available in {\bf every} video. Finally, it is important to state that transcriptions are not punctuated by YouTube. We use ML models to correct this behavior on the YouTube dataset, as well as the CORAAL/Buckeye dataset (below).

Considering the observational nature of our study, an extensive effort was applied to control the confounding variables' effect on the conclusions. The selection of the most prestigious movies of all time was an attempt to reduce the chance of having negative reviews, which would comprise higher scores in the toxicity analysis. We also tried to find at least one single-speaker video review for every movie to reduce any sampling bias impact. More importantly, the reviews' first-person nature helps eliminate the possibility of other people's opinions influencing the argumentative paths. We also believe that a wider variety of content producers within a given demographic group reduces the influence of a single person on the conclusions. 

{\em On author inferred demographic variables: } It is worth noticing that identifying race/gender is subjective and prone to errors -- we only have our view and not the content producer's identification. Thus, we avoid using YouTube demographic variables as input in {\bf any} analysis. We employed author-inferred demographics to collect a diverse (to the extent possible) dataset of utterances. Instead, we rely on the \# of AAE Expressions as an explanatory variable. Nevertheless, we do see an increase in AAE expressions based on the inferred demographic on YouTube (see Table~\ref{tab:data}).

{\bf CORAAL and Buckeye:} The Corpus of Regional African American Language \cite{kendall2021corpus} is a long-term corpus developed and maintained by the University of Oregon with the support of the National Science Foundation. The dataset comprises more than 150 socio-linguistic interviews with African-American English speakers born between 1891 and 2005. The dataset contains the orthographic transcriptions of interviews, together with the person's age, gender, and city they live in. Thus, each interview from the corpus encompasses many subjects from a given city/community.

Unlike the YouTube data, the transcriptions here represent the entire sentence, accounting for complete punctuation, line-level notes, and even non-linguistic sounds. Beyond that, the data also tracks the interviewer's voice in the dialog. The interviews allow the speakers to talk freely about different topics, an interesting feature that emulates the diversity of daily interactions and mood variations. The dataset aggregates five major sub-corpora from different locations in the United States of America, namely, Atlanta (2017), Washington (1968 and 2016), Lower East Side (2009), Princeville (2004), Rochester (2016), and Valdosta (2017). 

The Buckeye \cite{pitt2005buckeye} corpus is an effort started in 1999 and supported by the National Institute on Deafness and Other Communication Disorders and the Office of Research at Ohio State University. The initial goal was to gather approximately $300,000$ words of speech conversation from central Ohio speakers, keeping track of time and phonetic information. To reach that objective, researchers selected a group of 40 middle-class Caucasian speakers. 

Similar to the YouTube dataset, Buckeye sentences are not punctuated. However, instead of automatically generated captions, this corpus employed transcribers who were explicitly instructed not to use punctuation within the utterances and not to try to correct possible speech ``errors'' (we segmented sentences ourselves using an ML model; see below). 


\subsection{Data Segmentation and Cleaning}


Considering the observational nature of this study, we try to assess model distributions with as little influence of external variables as possible. To do so, we seek to find datasets capable of establishing a common setup for each of the demographics, i.e. a single speaker talking to an interviewer/audience, the ability of the speaker to talk freely, well-defined demographics, and similar informality standards. Experimenting with many models is also useful to attenuate the biases from a specific model and visualize more general trends.

\subsubsection{Sentence Segmentation}

Except for CORAAL, the datasets don't necessarily follow the correct orthographic rules about punctuation. Considering the other two transcripted corpus (i.e. YouTube and Buckeye), we should expect their sentences to be segmented not according to their inherent meaning but to silent intervals (not necessarily long ones) after a continuous pronunciation of words. Such segmentation can drastically misrepresent the sentences' meanings and consequently derive misleading conclusions about the data.

To reduce the impacts of incorrect segmentation in later analysis, we employed a machine learning-based segmentation to all corpus, except the one from Twitter. We believe tweets are self-contained messages where punctuation is not necessarily crucial to the audience's understanding. Consequently, segmentation is not necessary. On the other hand, we segment the only correctly punctuated corpus, CORAAL. Since we intend to compare the CORAAL dataset directly against Buckeye's, we should try to reduce the confounding factors (segmentation). The segmentation task was performed using NVIDIA's Punctuation and Capitalization model, available with the NeMo Toolkit\footnote{\url{https://github.com/NVIDIA/NeMo}}. 

\subsubsection{On the Impact of Swear Words} We executed two versions of our experiments, one considering utterances with swear words and another without. The swear words we considered were taken from the No Swearing project\footnote{\url{https://www.noswearing.com/}}, a cooperative effort to help programmers remove unwanted language from their applications. At the time of writing, there were 363 curse words listed by the project. Overall, we found no statistical difference in our results nor any significant difference in our figures. Thus, we decided to present our results with utterances as they are (without removing utterances with swear words).


\subsubsection{Linguistic and Grammatical Features}

One of the most relevant aspects of our analysis derives directly from controlling for linguistic and grammatical features from the available utterances. Thus, our research focuses on word classes, or Part-of-Speech (PoS), (e.g., Verb, Noun, Adjective, etc.) and language dimensions that represent psychological aspects of communication (e.g., Anger, Hate, Happiness, etc.).

The PoS features consider the function of each word in the sentence. The word \textit{smile} can be considered a verb; however, it can also be considered an adjective when used in certain scenarios, as in ``\textit{The smiling baby is really cute}''. This information can help us understand the sentence's composition regarding word classes. To classify the tokens according to their PoS categories, we employ a black-box model\footnote{\url{https://spacy.io}}.

To define linguistic features, we used the Linguistic Inquiry and Word Count (LIWC) software~\cite{pennebaker2001linguistic} in its 2015 release. LIWC is a research effort that maps words to psychological features (i.e., language dimensions) of speech. A single word may be assigned to as many suitable categories as necessary. For example, the word \textit{cried} is a 10-categories term (i.e., Affect, Positive Tone, Emotion, Negative Emotion, Sad Emotion, Verbs, Past Focus, Communication, Linguistic, and Cognition). The complete list of LIWC dimensions can be found in~\cite{boyd2022development}.


LIWC and PoS features and AAE expressions are computed based on the time each feature or PoS tag appears in a text. AAE expressions are based on the sum of occurrences of all expressions in an utterance.

\section{Results} \label{sec:results}

We now present our results. Initially, we compare the model scores for utterances with and without AAE expressions. After presenting the difference in score distributions, we use Logistic Regression models to determine which factors impact the outcome of different methods. Here, we constructed semantic and PoS tagging features from the utterances and information about demographics (except for YouTube) and the number of AAE expressions. Our final analysis uses recent language models~\cite{wang2020minilm} to compare sentence semantics. Statistically speaking, when pairs of utterances are very similar and have very divergent scores, a bias exists towards the utterance with more AAE expressions.

\subsection{Comparing Scores Per Usage of AAE Expressions}

\begin{figure}[t!]
\begin{subfigure}[b]{\textwidth}
         \centering
         \includegraphics[width=1\textwidth]{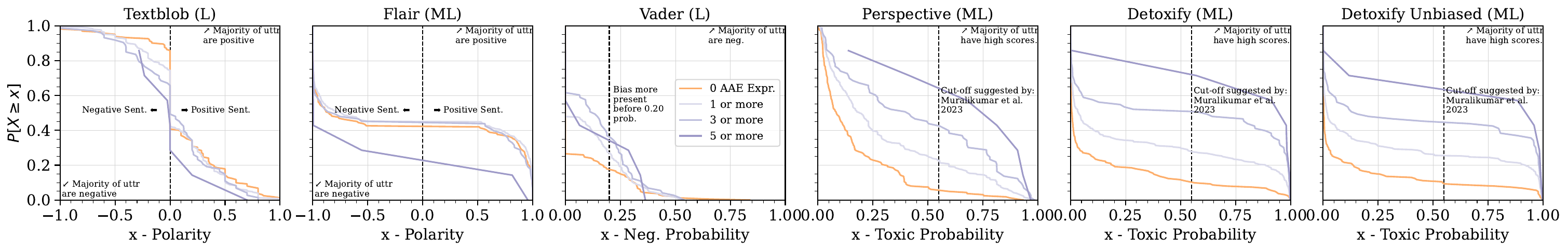}\vspace{-1em}
         \caption{Twitter}
         \label{img:twitter_w_wo_aae}
\end{subfigure}
\begin{subfigure}[b]{\textwidth}
         \centering
         \includegraphics[width=1\textwidth]{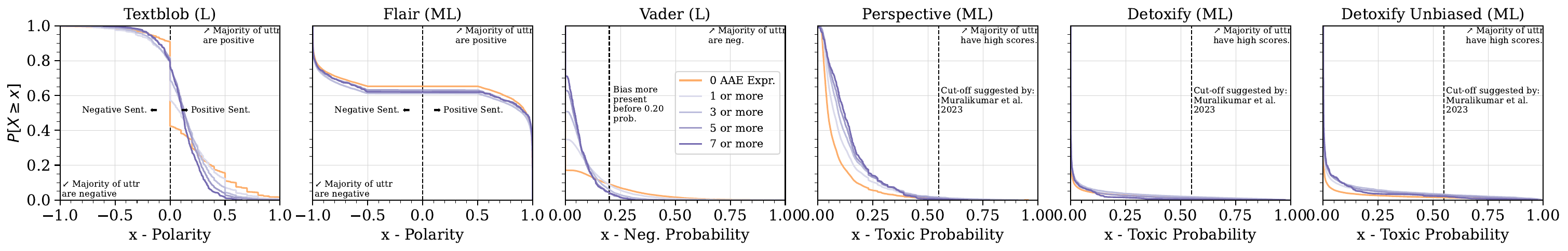}\vspace{-1em}
         \caption{YouTube}
         \label{img:youtube_w_wo_aae}
\end{subfigure}
\begin{subfigure}[b]{\textwidth}
         \centering
         \includegraphics[width=1\textwidth]{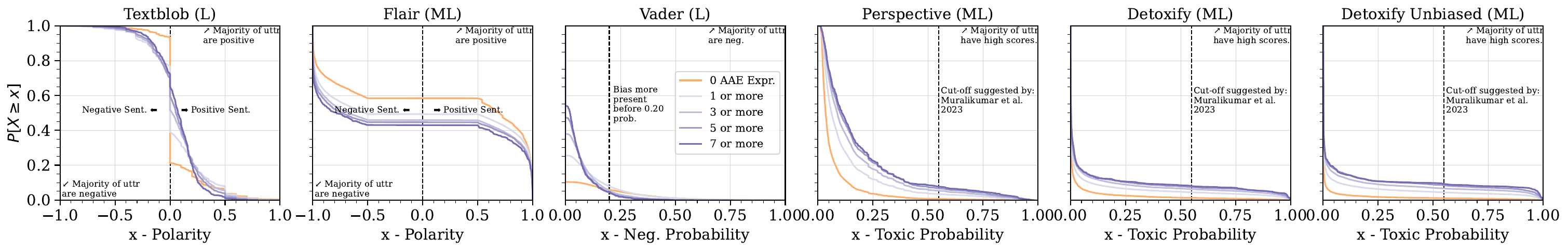}\vspace{-1em}
         \caption{CORAAL/Buckeye}
         \label{img:buckeye_vs_coraal_w_wo_aae}
\end{subfigure}
\caption{Score distributions for sentences with and without African American Expressions.}\vspace{-1em}
\end{figure}

Figure~\ref{img:twitter_w_wo_aae} compares toxicity/sentiment scores for utterances with and without AEE expressions. The figure shows the complementary cumulative distributions (CCDF) considering the number of AAE expressions on the utterance. That is, the x-axis of the figure shows the score, whereas the y-axis captures the fraction of sentences for that with scores greater than the one on the x-axis. In this and the other CDFs presented in our study, some care must be taken when interpreting results from Textblob and Vader. For these methods, negative values point toward negative sentiment, whereas positive values point toward positive sentiment. For the other models, 0 commonly indicates a not-negative (Vader) sentiment or non-toxic, whereas 1 is a negative sentiment or toxic utterance. This difference comes from sentiment analysis methods commonly measure {\em polarity} from -1 to 1. Unlike Textblob and Flair, Vader returns the probability of a negative, positive, or neutral score (adding up to one). Here, we focus on the negative probability. 

We can initially see that utterances with AAE expressions receive much higher scores for Twitter and toxicity models (Perspective, Detoxify, and Detoxify Unbiased). There is a clear tendency for statistical dominance -- the CDF, or y-value, of utterances with AAE expressions is above the one without AAE expressions regardless of x-values. This behavior is attenuated as more AAE expressions are considered in the utterance (we considered those with at least one, three, five, or seven expressions). For instance, considering Perspective on Twitter, the highest 20\% scoring (more toxic) utterances {\em without} AAE expressions achieve a minimum of 0.25. For utterances {\em with at least one} AAE expression, this minimum is around 0.55. A similar trend occurs on Detoxify and Detoxify Unbalanced.

Using a Kolmogorov-Smirnov test, we compared whether each CDF with AAE expressions differs from those without. Under $p<0.01$, Flair did not show this difference on the Twitter dataset when considering utterances with at least one expression. This is the {\bf only} case where we failed to reject the null hypothesis. When we consider lexical models, Textblob and Vader, it appears they present a mild bias on lower-scoring sentences. To help understand this issue, focus on the sentences with scores below 0.5 for Textblob and 0.20 for Vader. Whether this is an issue will depend on the cut-off developers employ; however, even a moderate cutoff of $\pm0.5$ appears to mitigate the issue. 


From Figures~\ref{img:youtube_w_wo_aae} and~\ref{img:buckeye_vs_coraal_w_wo_aae}, we can see the same trend that occurred on Twitter for the lexical approaches (Textblob and Vader), also occurs on YouTube and CORAAL/Buckeye. This finding likely stems from the fact that such approaches employ manually curated rules that do not consider AAE, a positive aspect of these approaches. On these datasets, Flair is biased, achieving lower scores (polarity or tendency to rate as more negative) for utterances with AAE expressions. The ML models still present biases on YouTube (Figure~\ref{img:youtube_w_wo_aae}). However, it is important to point out two facts: (1) albeit statistically significant, this bias is negligible on Detoxify for YouTube; (2) such bias is less present when considering the suggested cut-off of 0.55~\cite{muralikumar2023human} for toxicity models. This is not to say AAE expressions will not bias these models. However, this bias {\bf may not} be an issue when using large cut-off values.

Again, ML models present a higher degree of bias in spoken English interviews (Figure~\ref{img:buckeye_vs_coraal_w_wo_aae}). Here, except for Perspective, we can still see the statistical dominance that was present on Twitter. To further investigate biases, we present an analysis of which features are good predictors of scores (below).

\subsection{How do Grammatical and Linguistic Features impact Scores?}

\begin{table}[t!]
\centering
\scriptsize
\begin{tabular}{p{5cm}rr|rrrr}
\toprule
\textbf{Features} & \textbf{Textblob (L)} & \textbf{Flair (ML)} & \textbf{Vader (L)} & \textbf{Perspective (ML)} & \textbf{Detoxify (ML)} & \textbf{Detoxify U. (ML)} \\
& \multicolumn{2}{c|}{\colorbox{OrangeRed}{\faThumbsODown$\,<0$}, \colorbox{SpringGreen}{\faThumbsOUp$\,>0$}}& \multicolumn{4}{c}{\colorbox{OrangeRed}{\faThumbsODown$\,>0$}, \colorbox{SpringGreen}{\faThumbsOUp$\,<0$}}\\

\midrule
AAE\_EXPR &   &   & \cellcolor{OrangeRed}\textbf{0.0934} & \cellcolor{OrangeRed}0.2238 & \cellcolor{OrangeRed}0.1779 &   \\ \hdashline[0.5pt/5pt]
LIWC\_SWEAR &   &   &   & \cellcolor{OrangeRed}\textbf{0.8567} & \cellcolor{OrangeRed}\textbf{0.9492} & \cellcolor{OrangeRed}\textbf{1.2792} \\
LIWC\_SEXUAL & \cellcolor{OrangeRed}\textbf{-0.3354} & \cellcolor{OrangeRed}\textbf{-1.3448} &   & \cellcolor{OrangeRed}\textbf{0.4657} & \cellcolor{OrangeRed}\textbf{0.5942} & \cellcolor{OrangeRed}\textbf{0.5609} \\
LIWC\_NETSPEAK &   & \cellcolor{OrangeRed}\textbf{-2.6379} &   & \cellcolor{SpringGreen}\textbf{-0.4496} & \cellcolor{SpringGreen}\textbf{-0.8121} & \cellcolor{SpringGreen}\textbf{-0.9239} \\
LIWC\_INFORMAL &   & \cellcolor{SpringGreen}\textbf{2.5228} &   & \cellcolor{OrangeRed}\textbf{0.4386} & \cellcolor{OrangeRed}\textbf{0.7988} & \cellcolor{OrangeRed}\textbf{0.946} \\
LIWC\_NEGATE &   & \cellcolor{OrangeRed}\textbf{-0.8331} &   & \cellcolor{OrangeRed}0.2075 &   & \cellcolor{OrangeRed}0.1855 \\
LIWC\_FILLER &   & \cellcolor{OrangeRed}\textbf{-2.1683} &   &   & \cellcolor{SpringGreen}\textbf{-0.6053} & \cellcolor{SpringGreen}\textbf{-0.6377} \\ 
LIWC\_ASSENT &   &   &   & \cellcolor{SpringGreen}-0.1599 & \cellcolor{SpringGreen}-0.2614 & \cellcolor{SpringGreen}-0.1985 \\
LIWC\_MALE & \cellcolor{OrangeRed}\textbf{-0.3237} &   &   & \cellcolor{SpringGreen}-0.1505 & \cellcolor{SpringGreen}-0.1968 & \cellcolor{SpringGreen}-0.191 \\ \hdashline[0.5pt/5pt]
POS\_X & \cellcolor{SpringGreen}\textbf{0.5655} &   & \cellcolor{SpringGreen}\cellcolor{SpringGreen}\textbf{-0.1636} & \cellcolor{SpringGreen}\textbf{-0.3857} & \cellcolor{SpringGreen}-0.4255 & \cellcolor{SpringGreen}-0.3066 \\
POS\_DET &   &   &   & \cellcolor{OrangeRed}0.1623 &\cellcolor{OrangeRed} 0.3369 & \cellcolor{OrangeRed}0.3268 \\ \hdashline[0.5pt/5pt]
DEMOGRAPHIC &   &   &   & \cellcolor{OrangeRed}0.0508 & \cellcolor{OrangeRed}0.1262 & \cellcolor{OrangeRed}0.0762 \\\bottomrule
\end{tabular}
\caption{Logistic Regression coefficients for the Twitter dataset with $p < 0.05$. Each model's five most relevant coefficients are presented in bold, whereas not statistically significant coefficients were omitted. When a coefficient pushes towards a negative sentiment or toxic score, we color it \colorbox{OrangeRed}{red (\faThumbsODown)}. Positive sentiment and non-toxic score is colored \colorbox{SpringGreen}{green (\faThumbsOUp)}.\vspace{-2em}}
\label{table:lr_twitter}
\end{table}

In Table~\ref{table:lr_twitter}, we present our Logistic regression results for Twitter. For each tweet, we counted the number of AAE expressions, LIWC categories, and PoS tags as features. Being counts, all of these features have positive values only. To present regression coefficients on a similar, features were {\em Min-Max} scaled to the $[0, 1]$ range before the regression was executed. Models were executed with an intercept variable and no regularization. Models that output polarity had such polarity values re-scaled to $[0, 1]$ also (by adding one and dividing by two). The table presents only the statistically significant features ($p < 0.05$). LIWC features are identified by \texttt{LIWC\_}, and PoS tags by \texttt{POS\_}. The demographic variable (AAE tweet or non-AAE tweet) was used as a Twitter and CORAAL/Buckeye feature. 

{\em Due to space constraints, we do not present regression tables for YouTube nor CORAAL/BUCKEYE}. Results for these datasets are discussed throughout the text (these tables will be made available as supplementary material).

Our feature of most interest is the \texttt{AAE\_EXPR} (the number of AAE expressions on the utterance). The other features act as control variables to ensure that, on some level, such expressions are not being confounded with other grammatical/linguistic attributes of the sentence.

From the table, we can see that this feature pushes the polarity of the Vader model towards having fewer sentiments. The feature is not significant for the other sentiment analysis models. Nevertheless, this feature is significant for Perspective and Detoxify (but not for Detoxify Unbiased). When we consider YouTube, \texttt{AAE\_EXPR} is statistically significant for Vader (\colorbox{OrangeRed}{0.0617}), Perspective (\colorbox{OrangeRed}{0.2488}), and -- {\em suprisingly} -- Detoxify Unbiased (\colorbox{OrangeRed}{0.1334}). On CORALL/Buckeye, it was significant for Flair (\colorbox{OrangeRed}{-0.9432}, the change in sign is expected for Flair, see Tables~\ref{tab:examples} and~\ref{table:lr_twitter}), Perspective (\colorbox{OrangeRed}{0.3291}), Detoxify (\colorbox{OrangeRed}{0.1754}) and -- {\em again suprisingly} -- Detoxify Unbiased (\colorbox{OrangeRed}{0.2126}).

For Twitter and CORAAL/Buckeye, our demographic variable (\texttt{DEMOGRAPHIC}) was used as a categorical feature. This feature was statistically significant for Twitter and not CORAAL/Buckeye. Twitter is the only dataset where the demographic variable was developed to align with AAE utterances. On CORAAL/Buckeye, an African-American may not employ AAE, or a Caucasian may employ AAE (in fact, the usage of expressions is comparable in Table~\ref{tab:data}).

Considering the other features, some linguistic features are expected to push models towards negative sentiment or toxic scores (this is the case for the feature \texttt{LIWC\_SWEAR} in every dataset). Finally, PoS features were only statistically significant for Twitter. This was the case for quantifiers, \texttt{POS\_DET}, and the unknown/other tag, \texttt{POS\_X}.

Overall, our results agree with our CDF comparisons before. Biases are present in lexical models but to a lesser extent than in ML models (lower or non-significant regression coefficients for AAE expression usage). 

\subsection{Semantic Comparison}

We now present our final results concerning the semantics of utterances. Our approach seeks to answer the following question: {\em How do highly semantically similar pairs of utterances that achieve diverging scores differ in their usage of AAE expressions?} Notice that we have two conditions here: (1) being similar in meaning and (2) achieving diverging scores. Such pairs of utterances are interesting because they control for confounding factors in semantics.

We employ a recent Bert-based large language model as our semantic feature extractor~\cite{wang2020minilm}. With this model, utterances are mapped to an embedding vector. For pairs of utterances, we compare these vectors using a cosine similarity (indicating whether the embeddings point in the same direction for the pair): -1 indicates completely dissimilar, 0 indicates a lack of relationship, and 1 indicates completely similar. We deem two sentences semantically similar when the cosine score is above 0.5 (less than 0.1\% of pairs of utterances as discussed below).

For each dataset and sentiment/toxicity method of our study, we isolated the top 2.5\% and bottom 2.5\% scoring utterances. To perform a single analysis, we standardized scores. For Perspective and Detoxify, the top 2.5\% have a higher chance of being toxic, whereas for Vader, Flair, and Textblob, the top 2.5\% have negative polarity. Due to memory constraints sampled 100,000 of such pairs per dataset and method. Next, we focused only on pairs where one of the sentences had at least one AAE expression. If this is not the case, the difference in score certainly is not due to the number of AAE expressions employed. When this is the case, the usage of AAE expressions may be the underlying cause. Due to the small sample size, we did not find any pairs on Twitter that met our conditions. 

Combining every method, we found 585,679 unique pairs on YouTube with diverging scores (our top 2.5\% versus the bottom 2.5\%). Out of these, 568 had a cosine similarity above 0.5\%. On CORAAL/Buckeye, we found 592,606 unique pairs from our diverging scores filter, with 243 being highly similar. For each setting (YouTube or CORAAL/Buckeyey), we computed the number of pairs where: ($d_{bias}$) the most toxic or most negative had more AAE expressions; ($d_{eq}$) both had the same amount of expressions, and ($d_{no\_bias}$) the least toxic or most positive had more AAE expressions. 

For YouTube, we have that: $d_{bias} = 272\,(48\%)$, $d_{eq} = 87\,(15\%)$, and $d_{no\_bias} = 209\,(37\%)$. Whereas on CORAAL/Buckeye: $d_{bias} = 187\,(77\%)$, $d_{eq} = 26\,(11\%)$, and $d_{no\_bias} = 30\,(12\%)$. Under a Binomial test and $p<0.01$, in both cases, we find that $d_{bias} > d_{no\_bias}$. Thus, results are likely not due to random chance, showing statistical evidence of bias\footnote{$d_{eq}$ is not considered as the counts of AAE expressions usage are certainly not the issue in these settings (they are equal)}.


\section{Conclusions and Limitations} \label{sec:conc}

This paper investigates the capability of sentiment analysis/toxicity methods to skillfully disambiguate harmful situations and normal events regarding the usage of AAE expressions. We analyzed the performance of six well-known off-the-shelf methods in light of four different datasets. Our datasets ranged from online texts from Twitter, single-speaker closed captions from YouTube, and spoken English encompassing daily live situations. Overall, our analyses performed broadly to isolate confounding variables from our main focus, the usage of African-American English expressions.

Considering the latter, or nonexistent, introduction of AAE in machine learning datasets, the under-representation of such expressions as non-toxic (depending on the situation) leads ML models to present a systemic bias towards AAE. We argue that the biggest problems derive directly from the absence of context in the utterances. Since they employ human-crafted rules, lexical-based (rule) approaches tend to be less biased than ML models.

Similarly to any observational study, our results are impacted by the data sampling strategy. We present a broad-scale analysis covering several datasets/methods to mitigate this. We also employed linguistic and grammatical features as control attributes to isolate the biased effect of AAE expressions and language model distance analyses, controlling for semantics. We studied several methods and found similar findings in \textbf{virtually all} methods and datasets with different effect sizes, giving us confidence about our key findings. 

Thus, considering the above arguments, our study shows that biases towards AAE expressions\footnote{\url{https://anonymous.4open.science/r/aae_bias-D396/data/aae_terms_black_talk.yaml}} are systemic, meaning that all evaluated models presented some level of bias. Thus, we hope our observations might provoke the incorporation of AAE expressions, such as the ones we make available, in training datasets to mitigate such biases.



\section*{Ethical Considerations}

We now present our ethical considerations. 

{\bf Ethical Concerns:} One of the ethical concerns of our study comes from using demographic variables related to race {\bf without} ground-truth self-identification labels for speakers. To mitigate this issue, we refrained from using author-inferred demographic variables in our study (on the YouTube dataset). CORAAL/Buckeye are well-established in linguistics, with CORAAL focusing solely on African Americans. This issue is not present on Twitter, as labels come from using AAE or not (regardless of race). We also point out that our main statistical variable of study is not race. We focus on the usage of AAE expressions, where such expressions came from reliable and suggested sources (by the organizers of a well-known dictionary).

{\bf Unintended Impact:} Readers may interpret our research as against ML models or automatic utterance scoring tools. We point out that this is {\bf not} our statement. Our research advances both recent and large literature on the unintended biases of Lexical/AI/ML models. We hope our findings will improve how such tools are used; model advances towards fewer biases or both.

{\bf Researcher Background:} The majority of authors of this study are from a region where racial discrimination is still very present in the population's day-to-day lives. Our research aims to foster the ongoing discussion on how AI impacts the lives of different historically segregated communities. 

If readers deem any terms or expressions used in this paper offensive, we point out that it was not deliberate.


\bibliographystyle{ACM-Reference-Format}
\bibliography{bibs}

\end{document}